\documentclass{article}

\usepackage[preprint]{neurips_2025}

\usepackage[utf8]{inputenc}
\usepackage[T1]{fontenc}
\usepackage{hyperref}
\usepackage{url}
\usepackage{booktabs}
\usepackage{amsfonts}
\usepackage{nicefrac}
\usepackage{microtype}
\usepackage{xcolor}
\usepackage{soul}
\usepackage{amsmath,amssymb,graphicx,color,algorithm,subfig, algpseudocode,booktabs,bm,relsize,enumitem,multirow,amsthm,epsfig,caption}
\usepackage[export]{adjustbox}
\usepackage{lscape}
\usepackage{multirow}
\usepackage{graphicx}
\usepackage{booktabs}
\usepackage{makecell}
\usepackage{duckuments}

\usepackage{microtype}
\usepackage{graphicx}
\usepackage{hyperref}

\usepackage{amsmath}
\usepackage{amssymb}
\usepackage{mathtools}
\usepackage{amsthm}
\usepackage{wrapfig}

\usepackage[capitalise,noabbrev,nameinlink]{cleveref}
\creflabelformat{equation}{#2\textup{#1}#3}
\creflabelformat{section}{#2\textup{#1}#3}
\creflabelformat{appendix}{#2\textup{#1}#3}
\creflabelformat{figure}{#2\textup{#1}#3}
\crefname{algocf}{Algorithm}{Algorithms}
\Crefname{algocf}{Algorithm}{Algorithms}

\definecolor{DarkBlue}{rgb}{0,0.08,0.45}
\definecolor{PlotBlue}{RGB}{1, 132, 234}

\hypersetup{
colorlinks=true,
linkcolor=DarkBlue,
citecolor=DarkBlue,
filecolor=DarkBlue,
urlcolor=DarkBlue}

\usepackage{amsmath,amssymb,graphicx,color,algorithm,subfig, algpseudocode,booktabs,bm,relsize,enumitem,multirow,amsthm,epsfig,caption}
\usepackage[export]{adjustbox}
\usepackage{lscape}
\usepackage{multirow}
\usepackage{graphicx}
\usepackage{booktabs}
\usepackage{makecell}

\newsavebox{\measurebox}

\definecolor{cornellred}{rgb}{0.7, 0.11, 0.11}
\definecolor{cadmiumgreen}{rgb}{0.0, 0.42, 0.24}
\definecolor{aliceblue}{rgb}{0.91, 0.94, 0.97}
\definecolor{darkblue}{rgb}{0.83, 0.89, 0.97}
\definecolor{Red7}{rgb}{0.941, 0.243, 0.243}
\definecolor{Green7}{RGB}{55, 178, 77}
\definecolor{Blue9}{rgb}{0.098,0.3,0.9}

\usepackage{pifont}

\sethlcolor{aliceblue}

\title{Learning Sim-to-Real Humanoid Locomotion \\ in 15 Minutes}

\author{%
  \hspace{-0.225cm}Younggyo Seo$^{*}$ \; Carmelo Sferrazza$^{*}$ \; Juyue Chen \\ \textbf{Guanya Shi} \; \textbf{Rocky Duan} \; \textbf{Pieter Abbeel}\\[0.25cm]Amazon FAR (Frontier AI \& Robotics) 
}

\begin{document}

\maketitle

\vspace{-0.2in}
\begin{abstract}
Massively parallel simulation has reduced reinforcement learning (RL) training time for robots from days to minutes.
However, achieving fast and reliable sim-to-real RL for humanoid control remains difficult due to the challenges introduced by factors such as high dimensionality and domain randomization.
In this work, we introduce a simple and practical recipe based on off-policy RL algorithms, i.e., \textbf{FastSAC} and \textbf{FastTD3}, that enables rapid training of humanoid locomotion policies in just 15 minutes with a single RTX 4090 GPU.
Our simple recipe stabilizes off-policy RL algorithms at massive scale with thousands of parallel environments through carefully tuned design choices and minimalist reward functions.
We demonstrate rapid end-to-end learning of humanoid locomotion controllers on Unitree G1 and Booster T1 robots under strong domain randomization, e.g., randomized dynamics, rough terrain, and push perturbations, as well as fast training of whole-body human-motion tracking policies.
We provide videos and open-source implementation at: \url{https://younggyo.me/fastsac-humanoid}.
\end{abstract}

\begin{figure*}[h]
\centering
\vspace{-0.15in}
\includegraphics[width=1.0\linewidth]{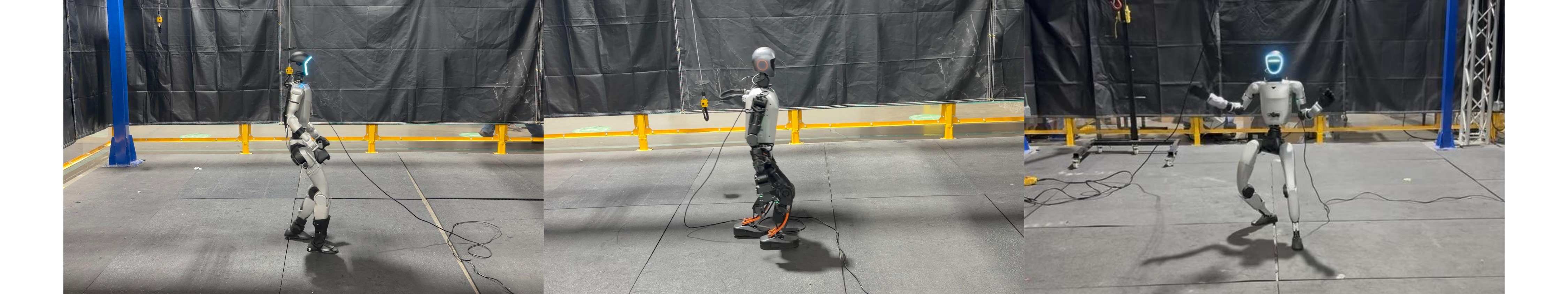}
\\
\includegraphics[width=1.0\linewidth]{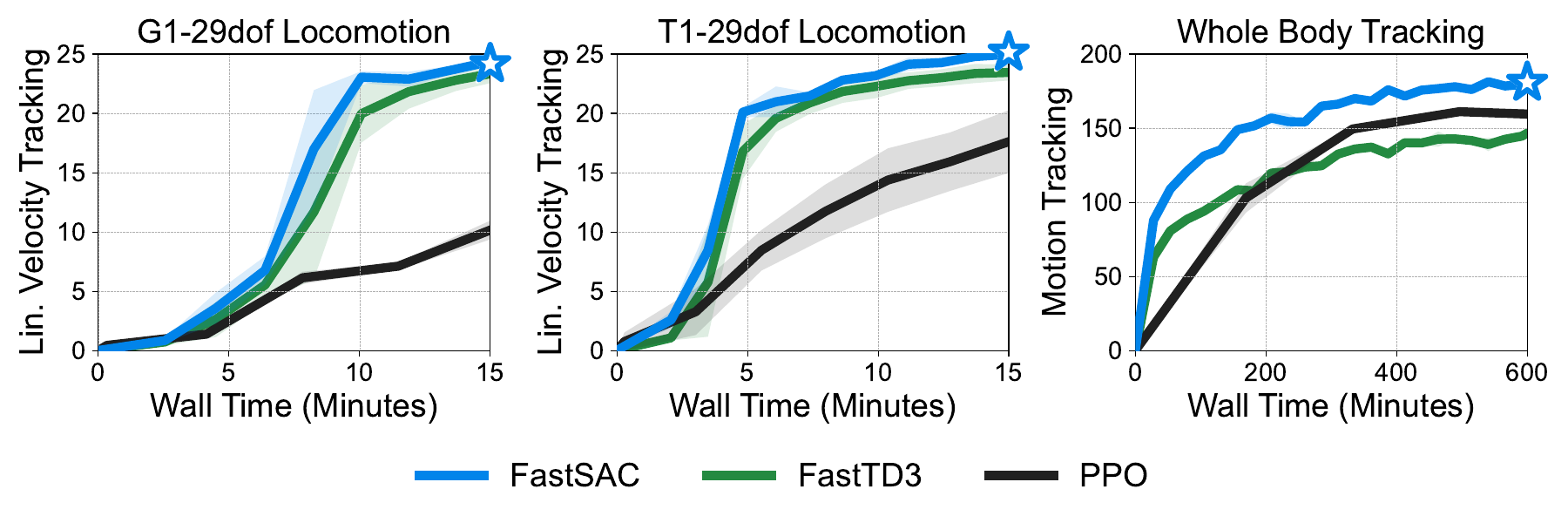}
\vspace{-0.25in}
\caption{\textbf{Summary of results.} We introduce a simple recipe based on off-policy RL algorithms, i.e., FastSAC and FastTD3, that learns robust humanoid locomotion policies in 15 minutes on a single RTX 4090 GPU, with strong domain randomization including randomized dynamics, rough terrain, and push perturbations.
We also show that our recipe based on off-policy RL algorithms is scalable and accelerates the training of whole-body tracking policies: trained with 4$\times$L40s GPUs and $16384$ parallel environments, FastSAC and FastTD3 learn to complete the full sequence of dancing motion much faster than PPO under the same condition.
For sim-to-real deployment with Unitree G1 and Booster T1, we used the checkpoints saved at the points we marked as \textcolor{PlotBlue}{$\bigstar$}.
}
\label{fig:aggregate_results}
\vspace{-0.1in}
\end{figure*}

\section{Introduction}

In recent years, reinforcement learning (RL) has undergone a dramatic shift driven by the emergence of massively parallel simulation frameworks \citep{rudin2022learning,kaufmann2023champion}.
By scaling environment throughput to thousands of environments, these frameworks have reduced wall-clock training time from many hours to mere minutes for a wide range of benchmark tasks \citep{makoviychuk2021isaac,mittal2023orbit,zakka2025mujoco}.
This shift has had an outsized impact on robotics, where sim-to-real development is inherently iterative: a policy is trained in simulation, deployed on hardware, and reveals mismatches such as unmodeled dynamics or sensing inaccuracies \citep{zhao2020sim}.
These discrepancies must then be corrected by improving the simulation environment, requiring the entire pipeline to be retrained \citep{chebotar2019closing}.
Because these cycles repeat until the policy is reliable, fast simulation becomes essential for making such iteration feasible.

Despite the speed offered by modern parallel simulators, these iterative cycles remain expensive in practice, especially for high-dimensional systems such as humanoids. Achieving robust transfer of policies to the real world typically requires expanding domain randomization~\citep{sadeghi2016cad2rl,tobin2017domain,peng2018sim}, randomizing terrain properties \citep{rudin2022learning}, or shaping curricula that encourage low-effort, stable whole-body behavior.
Such components complicate exploration and reduce sample efficiency, pushing training for humanoid locomotion or tracking back into the multi-hour regime.
Thus, despite dramatic gains in raw throughput, achieving fast, reliable sim-to-real iteration for humanoid control remains a challenge.

This work introduces a simple and practical recipe that brings sim-to-real iteration time for humanoid robots back to the order of minutes.
At the core of this recipe are FastSAC and FastTD3 \citep{seo2025fasttd3}, efficient variants of popular off-policy RL algorithms, i.e., Soft Actor-Critic \citep{haarnoja2018learning} and TD3 \citep{fujimoto2018addressing}, which have been shown to learn humanoid control policies faster than on-policy RL algorithms such as PPO \citep{schulman2017proximal}.
While \citet{seo2025fasttd3} demonstrated the first sim-to-real deployment of FastTD3 policies to real humanoid hardware, the results were limited to relatively simple controllers for humanoids with only a subset of joints.
In this work, we show that with careful design choices and hyperparameters, FastSAC and FastTD3 can scale to full-body humanoid control, enabling rapid sim-to-real iterations for training locomotion policies with all joints or whole-body tracking policies that follow human motion. 

Another important aspect of our fast sim-to-real recipe is its simplicity in reward design.
By adopting reward functions with essential terms, we can quickly sweep hyperparameters, isolate what matters for transfer, and avoid the brittle engineering often required in humanoid locomotion setups.
With our recipe, we train a full-fledged humanoid locomotion policy with randomized dynamics, rough terrain, push perturbations, and an automatic action-rate curriculum, all end-to-end in 15 minutes on a single RTX 4090 GPU.
The code for this recipe is available in the Holosoma repository \citep{Holosoma} at \url{https://github.com/amazon-far/holosoma}.

\section{Recipe}
\label{sec:recipe}

\subsection{FastSAC and FastTD3: Off-Policy RL for Humanoid Control}
\label{sec:recipe-off-policy-rl}

Our recipe is based on off-policy RL algorithms tuned for large-scale training with massively parallel simulation, i.e., FastTD3 and FastSAC \citep{seo2025fasttd3}, instead of PPO \citep{schulman2017proximal} that has been a standard algorithm for sim-to-real RL due to the ease of scaling up with parallel simulation.
This is motivated by recent work that have demonstrated off-policy algorithms can also scale effectively and be faster than PPO in various benchmark tasks \citep{li2023parallel,raffin2025isaacsim,seo2025fasttd3,shukla2025fastsac} by effectively re-using the data from simulation.
Notably, \citet{seo2025fasttd3} report the first sim-to-real deployment of humanoid control policies trained with off-policy RL to real humanoid hardware.
However, its results were limited to humanoid robots with a subset of joints.

This section describes how we train FastTD3 and FastSAC to achieve full-body humanoid control, enabling rapid sim-to-real iterations for training locomotion policies with all joints or whole-body tracking policies that follow human motion.
In particular, we improve the recipe of \citet{seo2025fasttd3} in training FastSAC, which learns how to explore environments from data with its maximum entropy learning scheme instead of deterministic policies with fixed noise schedules (FastTD3), mitigating the exploration challenges of training with strong domain randomization.

\begin{figure*}[t!]
\centering
\subfloat[Clipped double Q-learning]
{
\includegraphics[width=0.3125\linewidth]{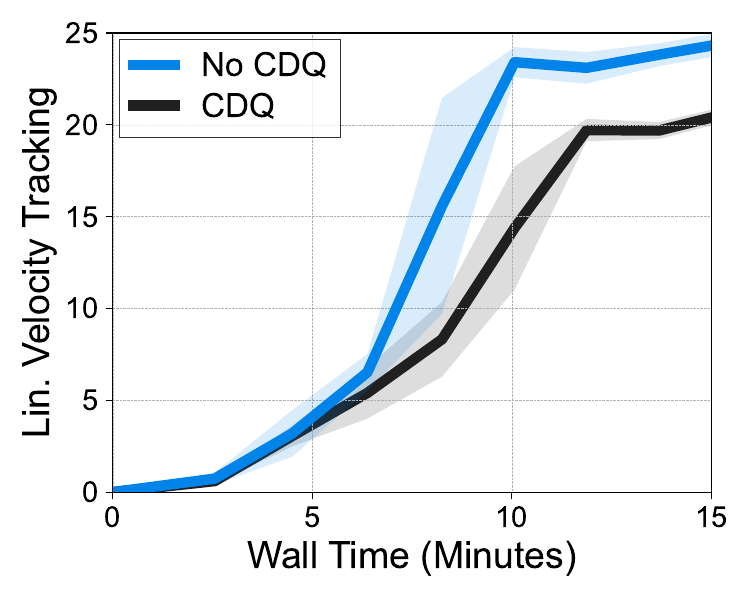}
\label{fig:fig1}
}
\subfloat[Number of update steps]
{
\includegraphics[width=0.3125\linewidth]{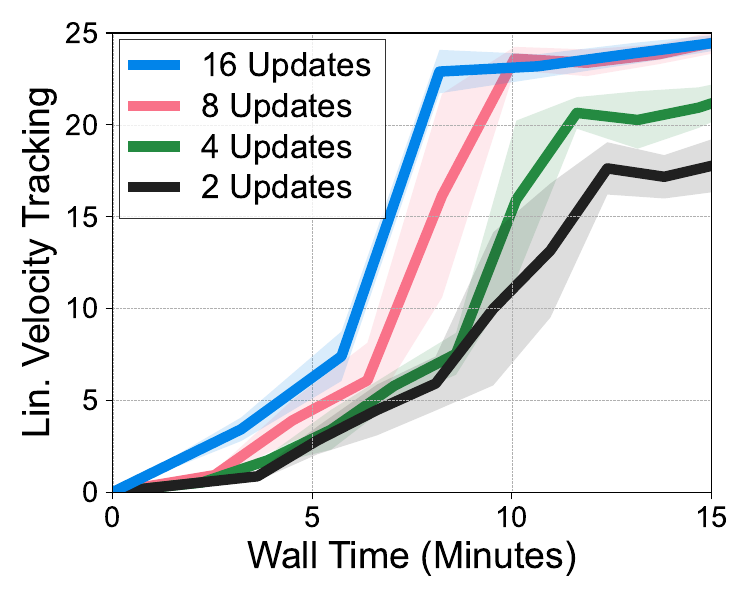}
\label{fig:fig2}
 }
\subfloat[Normalization]
{
\includegraphics[width=0.3125\linewidth]{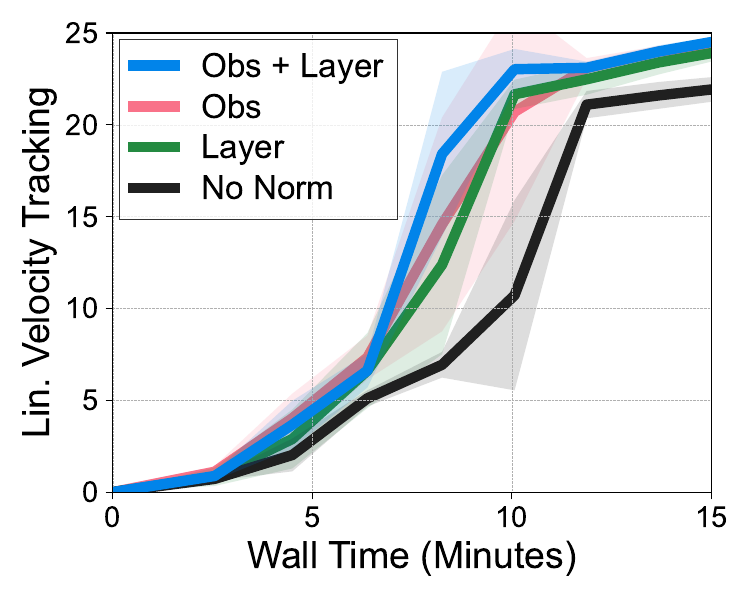}
\label{fig:fig3}
}\vspace{-0.15in}
\\ \subfloat[Effect of $\gamma$]
{
\includegraphics[width=0.3125\linewidth]{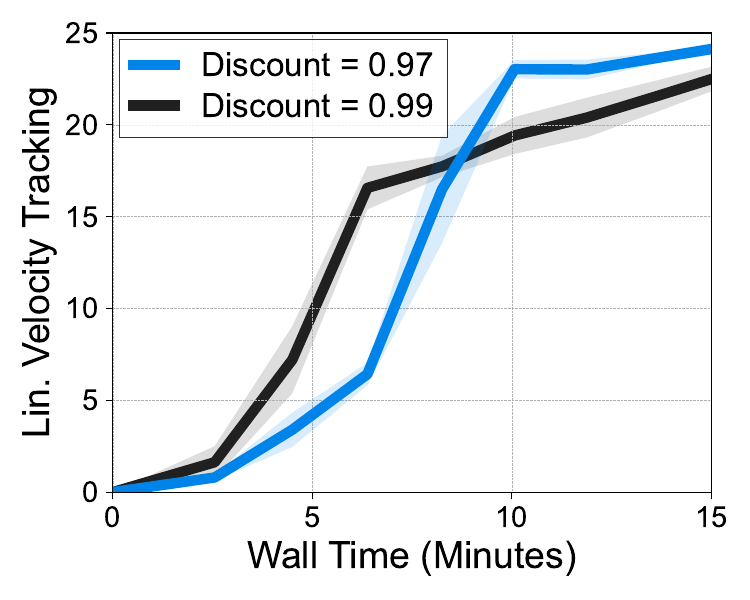}
\label{fig:fig4}
}
\subfloat[Effect of $\gamma$ (WBT)]
{
\includegraphics[width=0.3125\linewidth]{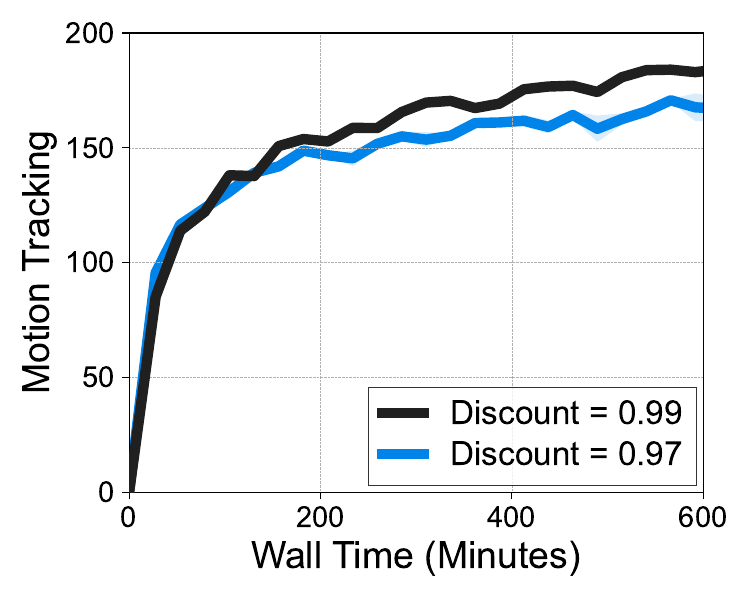}
\label{fig:fig5}
}
\subfloat[Number of environments (WBT)]
{
\includegraphics[width=0.3125\linewidth]{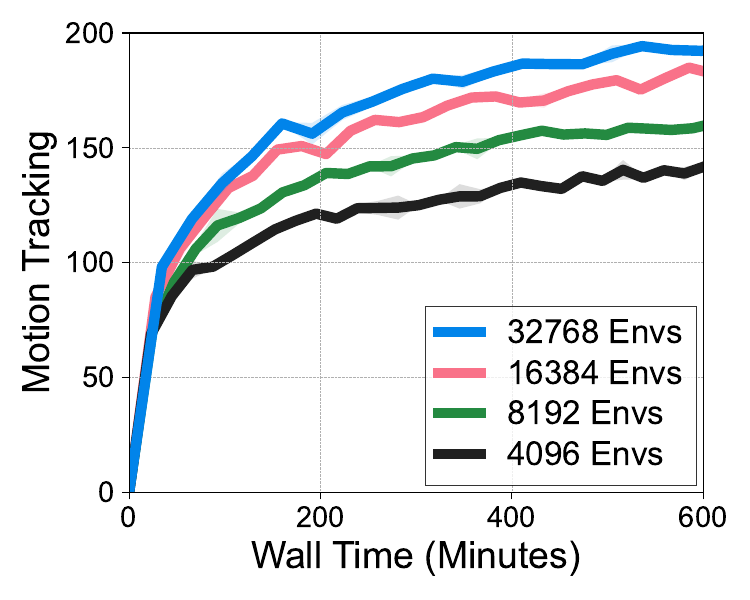}
\label{fig:fig6}
}
\caption{\textbf{FastSAC: Analyses.} We investigate the effect of (a) Clipped double Q-learning, (b) number of update steps, (c) normalization techniques, and (d) discount factor $\gamma$ on a Unitree G1 locomotion task with rough terrain.
We further investigate the effect of (e) discount factor $\gamma$ and (f) number of environments on a G1 whole-body tracking (WBT) task with a dancing motion.
We use a single RTX~4090~GPU for locomotion experiments (a-d) and 4$\times$L40s GPUs for whole-body tracking (e-f).
}
\label{fig:analysis}
\end{figure*}

\paragraph{Scaling up off-policy RL with massively parallel simulation}
Similar to prior work \citep{li2023parallel,shukla2025fastsac,raffin2025isaacsim,seo2025fasttd3}, we use massively parallel simulation for training FastSAC and FastTD3 agents.
We find that the effect of using more environments is particularly visible in challenging whole-body tracking tasks (see \cref{fig:fig6}).
We also find that most of observations in \citet{seo2025fasttd3} with regard to scaling up off-policy RL also holds for full-body humanoid control as well.
For instance, we find that using large batch size up to 8K consistently improves performance.
We also find that taking more gradient steps per each simulation step usually ends up in a faster training, and slow simulation speed often becomes a bottleneck with more challenging setups such as training robots in non-flat terrains (see \cref{fig:fig2}).
This makes off-policy RL, which can re-use data from previous interactions instead of discarding it, a more attractive choice for fast training.

\paragraph{Joint-limit-aware action bounds}
One challenge in training off-policy RL algorithms such as SAC or TD3 is setting proper action bounds for its Tanh policy.
For instance, \citet{raffin2025isaacsim} observed that training often becomes unstable when trained in unbounded action space.
To address this challenge, we introduce a simple technique that sets the action bounds based on the robots' joint limits when using PD controllers. In particular, we calculate the difference between each joint's limit and its default position, then use it as an action bound for each joint.
We find that this effectively reduces the need to tune action bounds for training FastSAC and FastTD3\footnote{Interestingly, after we have fully stabilized the training with all the components, we find that we can achieve stable training of FastSAC and FastTD3 agents with an unbounded action space. Nonetheless, we still keep joint-limit-aware action bounds as we expect this scheme to be helpful in training off-policy RL agents for other robots or tasks. We recommend training agents in an unbounded action space when encountering a failure case where the robot is not generating enough torque due to restrictive action bounds.
}.

\paragraph{Observation and Layer normalization} Similar to \citet{seo2025fasttd3}, we find that observation normalization is helpful for training.
However, unlike \citet{seo2025fasttd3}, we find that layer normalization \citep{ba2016layer} is helpful in stabilizing the performance in high-dimensional tasks (see \cref{fig:fig3}).
This is aligned with prior observations that find layer normalization is helpful for training SAC \citep{ball2023efficient,nauman2024bigger} agents in challenging benchmark tasks.

\paragraph{Critic learning hyperparameters} We find that using the average of Q-values improves FastSAC and FastTD3 performance over using Clipped double Q-learning (CDQ; \citealt{fujimoto2018addressing}) that uses the minimum (see \cref{fig:fig1}).
This aligns with the observation of \citet{nauman2024bigger} that shows CDQ is harmful when used with layer normalization.
We find that low discount factor $\gamma=0.97$ is helpful for simple velocity tracking tasks (see \cref{fig:fig4}), while $\gamma=0.99$ is helpful for challenging whole-body tracking tasks (see \cref{fig:fig5}).
Following prior work \citep{li2023parallel,seo2025fasttd3}, we also use a distributional critic, i.e., C51 \citep{bellemare2017distributional}.
We find that distributional critic with quantile regression \citep{dabney2018distributional} is too expensive in particular with large batch training.

\paragraph{FastSAC: Exploration hyperparameters} 
A widely-used implementation of SAC bounds the standard deviation $\sigma$ of the pre-tanh actions to be $e^{2}$ \citep{huang2022cleanrl}.
However, we find that, when combined with large initial value of the temperature $\alpha$, this sometimes causes instability due to excessive exploration.
We instead set the maximum $\sigma$ to be $1.0$ and initialize $\alpha$ with the low value of $0.001$.
We also find that using auto-tuning for maximum entropy learning \citep{haarnoja2018soft2} consistently outperforms using the fixed alpha values. For the target entropy, we find that using $0.0$ (for locomotion tasks) or $-|\mathcal{A}|/2$ (for whole-body tracking tasks) works best in practice.

\paragraph{FastTD3: Exploration hyperparameters}
Following prior work \citep{li2023parallel,seo2025fasttd3}, we use mixed noise schedule that randomly samples Gaussian noise standard deviation from the range $[\sigma_{\texttt{min}}, \sigma_{\texttt{max}}]$.
We find that using low values, i.e., $(\sigma_{\texttt{min}}, \sigma_{\texttt{max}}) = (0.01, 0.05)$, performs the best.

\begin{figure*}[t!]
\centering
\includegraphics[width=1.0\linewidth]{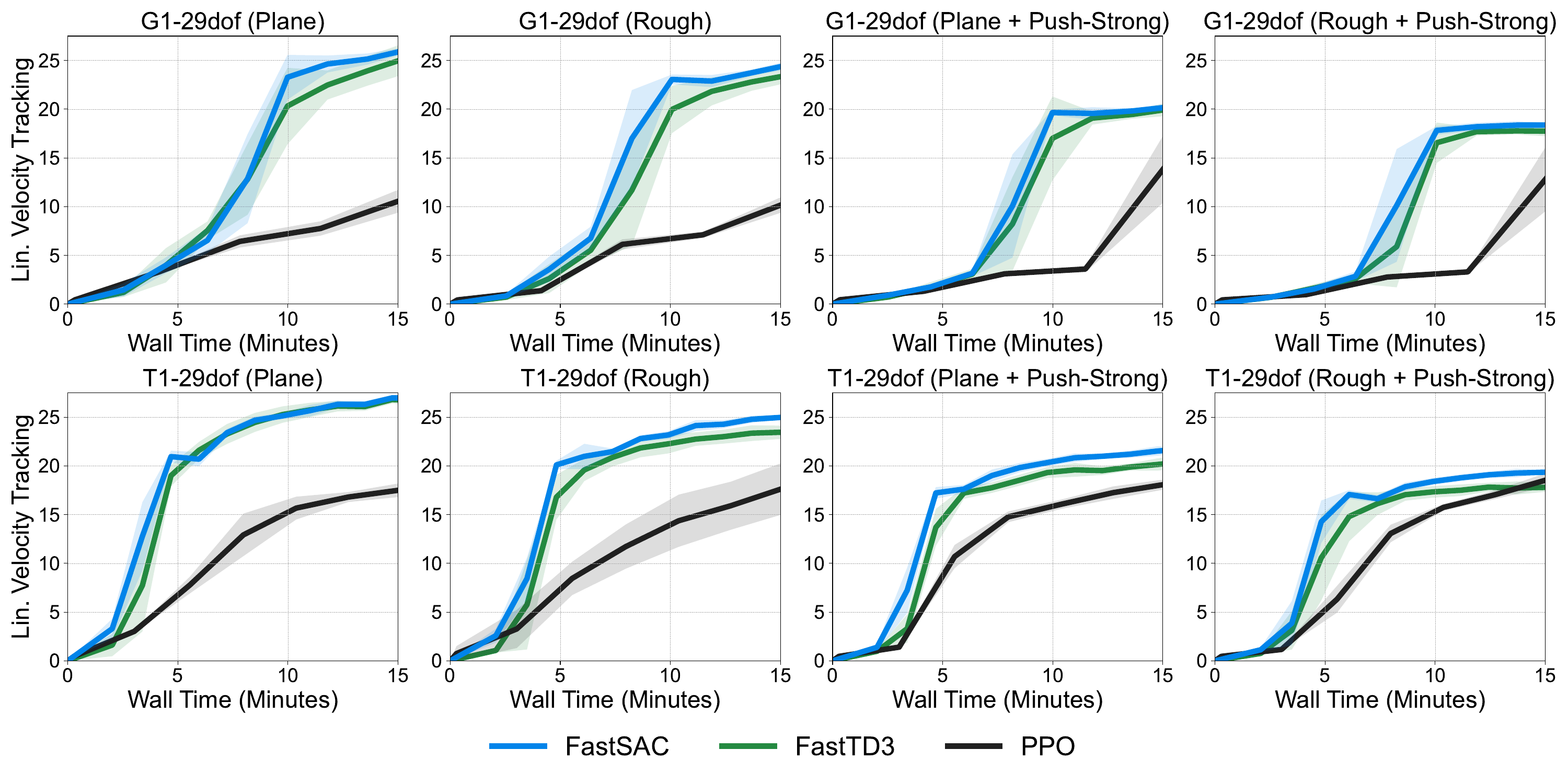}
\caption{\textbf{Locomotion (velocity tracking) results.} FastSAC and FastTD3 enable fast training of G1 and T1 humanoid locomotion policies with strong domain randomization such as rough terrain or $\texttt{Push-Strong}$ that applies push perturbations to humanoid robots every 1 to 3 seconds (max episode length is 20 seconds).
For non-$\texttt{Push-Strong}$ tasks, we apply push perturbations every 5 to 10 seconds.
We use a single RTX 4090 GPU for all locomotion experiments.
}
\label{fig:main_locomotion}
\end{figure*}

\paragraph{Optimization hyperparameters}
We train FastSAC and FastTD3 using Adam optimizer \citep{kingma2014adam} with a learning rate of $0.0003$.
We find that weight decay $0.1$, which \citet{seo2025fasttd3} uses, is a too strong regularization for high-dimensional control tasks and thus we use weight decay of $0.001$.
Similar to \citet{zhai2023sigmoid} where using low $\beta_{2}$ for Adam makes training stable with large batch sizes, we find that using $\beta_{2}=0.95$ slightly improves stability compared to using $\beta_{2}=0.99$.

\paragraph{Remark on additional techniques} We expect that recent advances in improving off-policy RL \citep{d'oro2023sampleefficient,schwarzer2023bigger,nauman2024bigger,lee2024simba,sukhija2024maxinforl,lee2025hyperspherical,obando2025simplicial} will be helpful for further improving the performance and stability of FastSAC and FastTD3. 
However, this work aims to keep the recipe as simple as possible and we expect the research community to advance the state-of-the-art based on our recipe.

\subsection{Simple Reward Design}
\label{sec:recipe-simple-reward-design}
Reward design for humanoid locomotion and whole-body control has traditionally depended on heavy reward shaping, often 20+ terms \citep{mittal2023orbit,HumanoidVerse}, e.g., tracking rewards for kinematic quantities, detailed posture regularizer, penalties on joint configurations, foot placement constraints, and shaping terms that strictly prescribe how the robot should move.
This complexity makes hyperparameter tuning difficult and often leads to brittle policy optimization.

Inspired by recent works that rely on much simpler reward functions \citep{zakka2025mujoco, liao2025beyondmimic}, we show that robust and natural behaviors can emerge from substantially simpler objectives (less than 10 terms).
Specifically, we adopt a minimalist reward philosophy that only adds a reward term if necessarily needed, and aim to have a nearly identical set of rewards across algorithms and robots.
Our goal is not to enforce a particular style, but to provide enough structure for robust locomotion and whole-body control while preserving behavioral richness.
Fewer reward terms also simplify hyperparameter tuning, enabling rapid sweeps crucial for sim-to-real iteration.

\paragraph{Locomotion (velocity tracking)}
We use a compact set of reward terms that cover only the essential components needed for stable humanoid gait transferrable from simulation to the real-world:
\begin{itemize}[topsep=1.0pt,itemsep=0.85pt,leftmargin=8mm]
    \item Linear and angular velocity tracking rewards to encourage the humanoid to follow commanded x-y speed and yaw rate. These are the main driver of emergent locomotion.
    \item A simple foot-height tracking term \citep{zakka2025mujoco,phase_rewards} to guide swing motion.
    \item A default-pose penalty to avoid extreme joint configurations.
    \item Feet penalties to encourage parallel relative orientation and prevent foot crossing.
    \item A per-step alive reward that encourages remaining in valid, non-fallen states.
    \item Penalties that keep the torso near a stable upright orientation.
    \item A penalty on the action rate to smooth control outputs.
\end{itemize}

We terminate the episode on ground contact by the torso or other non-foot body parts.
We also use symmetry augmentation \citep{mittal2024symmetry} to encourage symmetric walking pattern, which we also find to be helpful for faster convergence.
All penalties above are subject to a curriculum that ramp up their weights over the course of training as the episode length increases \citep{HumanoidVerse}, considerably simplifying exploration.
We find that these terms are sufficient to produce robust locomotion across rough terrain, with randomized dynamics and external perturbations, without relying on extensive reward shaping or other carefully tuned heuristics, and are applicable across multiple robots (i.e., G1 and T1) and algorithms (i.e., FastSAC, FastTD3, and PPO).

\paragraph{Whole-body tracking}
For whole-body tracking, we follow the reward structure introduced in BeyondMimic \citep{liao2025beyondmimic}, which already adheres to the same minimalist principles.
These rewards are built around tracking goals with lightweight regularization, together with DeepMimic-style termination conditions \citep{peng2018deepmimic}.
We additionally find that introducing external disturbances in the form of velocity pushes further robustifies sim-to-real performance.

\begin{figure*}[t!]
\centering
\includegraphics[width=1.0\linewidth]{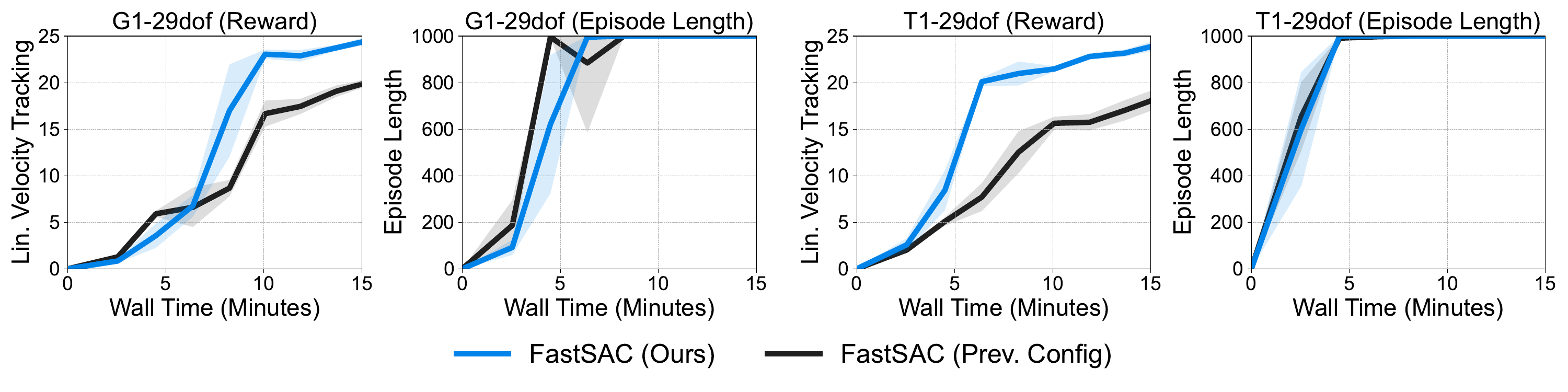}
\caption{\textbf{Improvement from our FastSAC recipe.}
While a version of FastSAC was previously considered as a baseline to FastTD3 \citep{seo2025fasttd3} in the context of humanoid control, a straightforward implementation of FastSAC exhibited training instabilities.
In this work, we have stabilized and improved FastSAC with a carefully tuned set of hyperparameters and design choices.}
\label{fig:main_comparison_to_prev_config}
\end{figure*}

\section{Experiments}

\subsection{Locomotion (Velocity Tracking)}

\paragraph{Setup}
For locomotion tasks, we train RL policies to maximize the sum of reward as we described in \cref{sec:recipe-simple-reward-design}, i.e., by training the robots to achieve the target linear and angular velocities while minimizing several penalty terms.
Throughout training, we randomly sample target velocity commands every 10 seconds.
When sampling the target commands, we randomly set the target velocities to zero with 20\% probability, so that the robot learns to stand instead of making it constantly walk on its position.
Unless otherwise specified, we train all robots on a mix of flat and rough terrains, which stabilizes robot walking in sim-to-real deployment.
We apply various domain randomization techniques to further robustify sim-to-real deployment: push perturbations, action delay, PD-gain randomization, mass randomization, friction randomization, and center of mass randomization (only for G1).
We report linear velocity tracking reward in all learning curves.

\paragraph{Results}
\cref{fig:main_locomotion} shows that FastSAC and FastTD3 quickly train G1 and T1 humanoid robots to track velocity commands in 15 minutes, significantly outperforming PPO in terms of wall-time clock.
We emphasize this is achieved in the existence of strong domain randomization: our humanoids learn to stand and walk in rough terrain, with consistent push perturbations, action delay, center of mass randomization, etc.
In particular, we observe that FastSAC and FastTD3 enables fast training of locomotion policies with strong domain randomization such as $\texttt{Push-Strong}$ that applies push perturbations to robots every 1 to 3 seconds, while PPO struggles with such strong perturbations.
We also find that FastSAC slightly outperforms FastTD3 in several locomotion setups, which we hypothesize to be due to efficient exploration through its maximum entropy exploration scheme.

\paragraph{FastSAC improvement over previous configuration}
\cref{fig:main_comparison_to_prev_config} shows how our recipe for training FastSAC improves the performance over the previous version of FastSAC trained with configuration from \citet{seo2025fasttd3}.
Specifically, we find that the use of layer normalization \citep{seo2025fasttd3}, disabling CDQ \citep{fujimoto2018addressing}, careful tuning of exploration and optimization hyperparameters is important for performance improvement (see \cref{sec:recipe} for details).

\begin{figure*}[t!]
\centering
\includegraphics[width=1.0\linewidth]{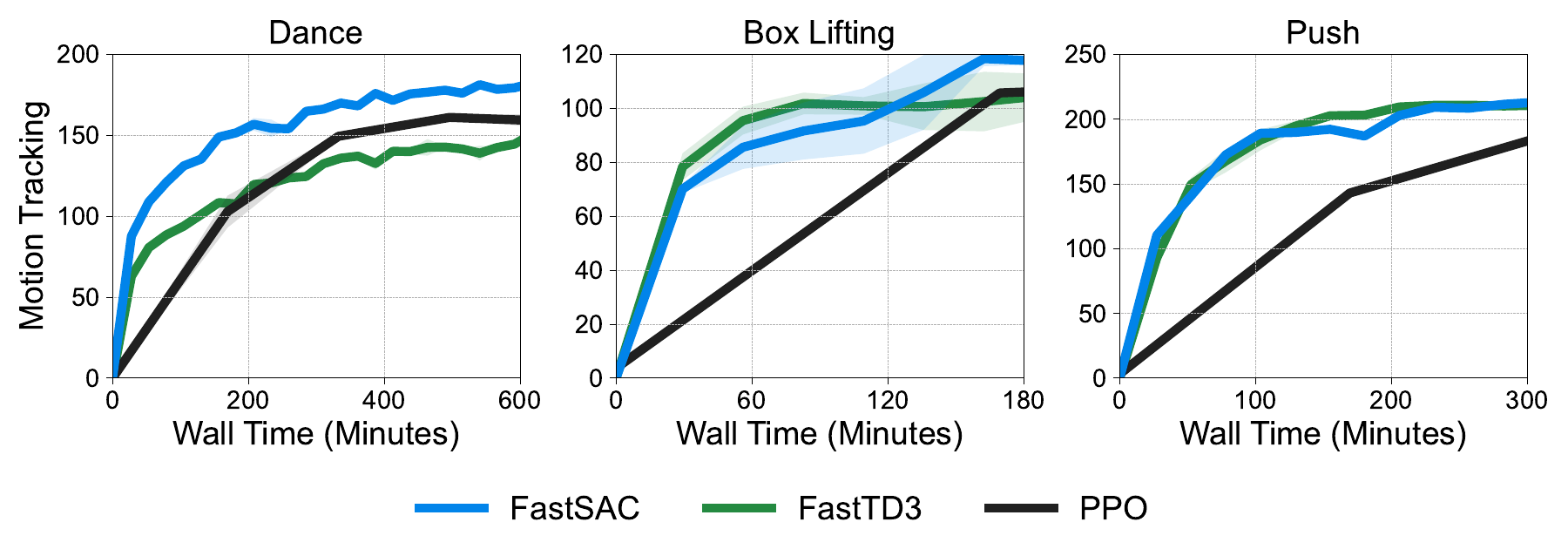}
\caption{\textbf{Whole-body tracking results.} We show that FastSAC and FastTD3 are competitive or superior to PPO in whole-body motion tracking tasks.
See \cref{fig:wbt_examples} for the sim-to-real deployment of FastSAC policies to real hardware.
We use 4$\times$L40s GPUs for all whole-body tracking experiments.
}
\label{fig:main_wbt}
\end{figure*}

\begin{figure*}[t!]
\includegraphics[width=1.0\linewidth]{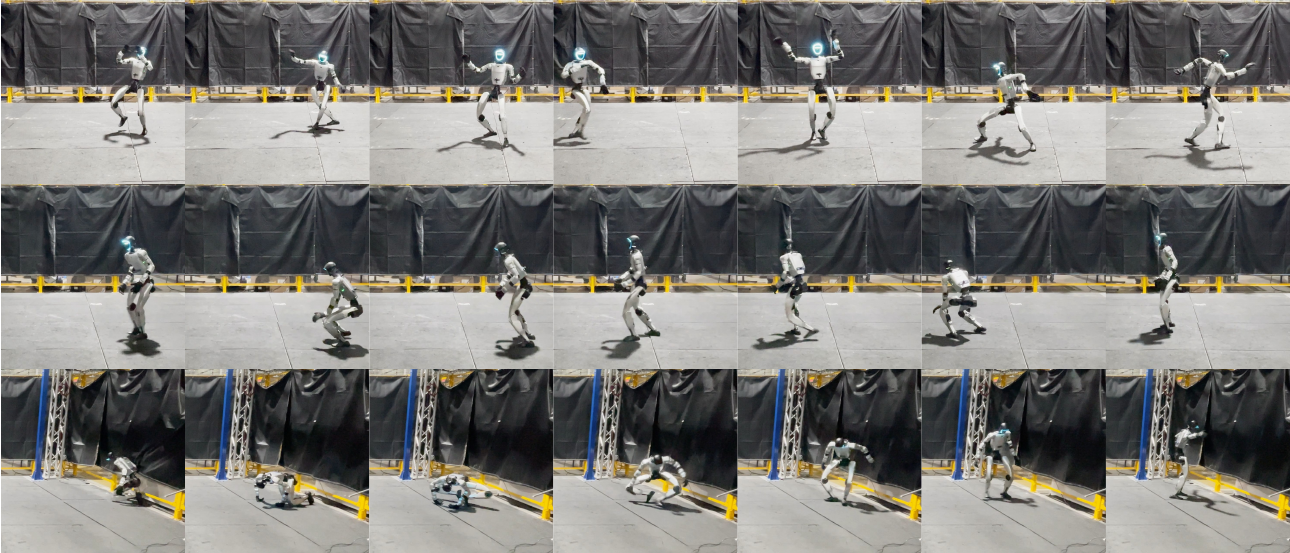}
\caption{\textbf{Whole-body tracking examples.} We demonstrate the sim-to-real deployment of whole-body tracking controllers for Unitree G1 trained with FastSAC (Top: Dance, Middle: Box Lifting, Bottom: Push).
Videos are available at \url{https://younggyo.me/fastsac-humanoid}.}
\label{fig:wbt_examples}
\end{figure*}

\subsection{Whole-Body Tracking}

\paragraph{Setup}
For whole-body tracking, we mostly follow the setup in BeyondMimic \citep{liao2025beyondmimic}.
We train RL policies to maximize the sum of rewards as we described in \cref{sec:recipe-simple-reward-design}.
We report the sum of tracking rewards in all learning curves.
Throughout training, we randomly sample motion segments for each episode.
Unlike BeyondMimic that minimizes the use of domain randomization for motion tracking, we find that using various domain randomization techniques stabilizes the behavior at deployment time.
In particular, we randomize friction, center of mass, joint position bias, body mass, PD gains, and also apply push perturbations.

\paragraph{Results}
\cref{fig:main_wbt} shows that FastSAC and FastTD3 can also quickly train G1 humanoid robots to track human motion, being competitive or superior to PPO.
We find that FastSAC outperforms FastTD3 in the $\texttt{Dance}$ task, which is a longer motion compared to the other tasks we considered.
We hypothesize better exploration through maximum entropy RL enables faster learning on more challenging tasks.
Further investigation into the performance difference between FastSAC and FastTD3 in more diverse types of tasks is an interesting future direction.

\paragraph{Sim-to-real deployment}
In \cref{fig:wbt_examples}, we further demonstrate the sim-to-real deployment of FastSAC whole-body tracking policies to real Unitree G1 humanoid hardware.
We find that FastSAC policies can complete several motions including the long motion like \texttt{Dance} that lasts more than 2 minutes.
These results show that FastSAC not only enables faster training in simulation but also actually learns deployable robust full-body humanoid control policies.

\section{Related Work}

\paragraph{Reinforcement learning with massively parallel simulation}
Massively parallel simulation has significantly reduced the wall-clock time required to train RL policies.
Early work primarily relied on CPU-based parallelization by launching simulations across multiple processes \citep{heess2017emergence,akkaya2019solving,stooke2018accelerated,espeholt2018impala,radosavovic2024learning}.
While policy learning often leveraged multiple GPUs, overall throughput remains bottlenecked by the process-management overhead and inherently slow simulation speed.
To address these limitations, the idea of using GPU-based parallel environments have been proposed \citep{liang2018gpu,makoviychuk2021isaac,mittal2023orbit,Genesis,zakka2025mujoco}, scaling up environment throughput to thousands of environments.
This has been the key driver of recent successes in training controllers for diverse robots with impressive capabilities \citep{rudin2022learning,agarwal2023legged,cheng2024extreme,singh2024dextrah,zhuang2024humanoid,li2025reinforcement,he2025hover,he2025asap}.
Building on this trend, our work focuses on accelerating sim-to-real iterations by combining massively parallel simulation with off-policy RL algorithms tuned for large-scale training regimes.

\paragraph{Algorithms for sim-to-real reinforcement learning}
Proximal policy optimization (PPO; \citealt{schulman2017proximal}) has been the de-facto standard algorithm for sim-to-real RL, and is often the only supported algorithm in widely used learning frameworks \citep{makoviychuk2021isaac,mittal2023orbit,zakka2025mujoco,schwarke2025rsl}, largely due to the ease of scaling up on-policy RL with massively parallel environments.
However, recent works have started to demonstrate that off-policy RL methods can also scale effectively in such large-scale training regimes \citep{li2023parallel,raffin2025isaacsim,shukla2025fastsac,seo2025fasttd3}.
Notably, \citet{seo2025fasttd3} report the first sim-to-real deployment of humanoid control policies trained with FastTD3, an efficient variant of TD3 \citep{fujimoto2018addressing} optimized for large-batch training with parallel simulation.
However, their results were limited to humanoid controllers only with a subset of joints.
In this work, we further push this direction by developing a sim-to-real RL recipe based on FastSAC and FastTD3 that achieve full-body humanoid control that controls all joints for locomotion or follows human motion.
While doing so, we have also stabilized and improved the performance of FastSAC, which has been shown to exhibit training instabilities for humanoid control in prior work \citep{seo2025fasttd3}, with careful design choices.

\renewcommand{\algorithmiccomment}[1]{\textsc{#1}}
\begin{algorithm*}[t]
\caption{{FastSAC: Pseudocode (distributional critic is omitted for simplicity)}}\label{alg:training}
\begin{algorithmic}[1]
\State Initialize actor $\pi_{\theta}$, two critics $Q_{\phi_{1}}, Q_{\phi_{2}}$, entropy temperature $\alpha$, replay buffer $\mathcal{B}$
\State Initialize target critics $Q_{\phi^{\texttt{target}}_{1}}, Q_{\phi^{\texttt{target}}_{2}}$ with $\phi^{\texttt{target}}_{1} \leftarrow \phi_{1}$ and $\phi^{\texttt{target}}_{2} \leftarrow \phi_{2}$
\For{each environment step} \vspace{0.01in}
\State Sample $a \sim \pi_{\theta}(o)$ given the current observation $o$, and take action $a$
\State Observe next state $o'$ and reward $r'$
\State Store transition $\tau = (o, a, o', r')$ in replay buffer $\mathcal{B} \leftarrow \mathcal{B} \cup \{\tau\}$
\For{$j=1$ {\bfseries to} $\texttt{num\_updates}$}
\State Sample mini-batch $B =\{\tau_{k}\}_{k=1}^{|B|}$ from $\mathcal{B}$
\State Compute target Q-value via average:
\State \;\; $y = r' + \dfrac{\gamma}{2}\displaystyle\sum_{i=1}^{2}\left(Q_{\phi^{\texttt{target}}_{i}}(o', \tilde{a}') - \alpha \log \pi_{\theta}(\tilde{a}'|o') \right)$ with $\tilde{a}' \sim \pi_{\theta}(\cdot|o')$
\State Update critic: 
\State \;\; $\phi_{i} \leftarrow \phi_{i} - \nabla_{\phi_{i}}\dfrac{1}{|
B|}\displaystyle\sum_{\tau_{k} \in B}\Big(Q_{\phi_{i}}(o, a) - y\Big)^{2}$ for $i \in \{1, 2\}$
\State Update actor with reparameterization trick:
\State \;\; $\theta \leftarrow \theta + \nabla_{\theta}\dfrac{1}{2|B|}\displaystyle\sum_{\tau_{k} \in B}\sum_{i=1}^{2}\Big(Q_{\phi_{i}}(o, \tilde{a}) - \alpha \log \pi_{\theta}(\tilde{a}|o)\Big)$ with $\tilde{a} \sim \pi_{\theta}(\cdot |o)$
\State Update entropy temperature:
\State \;\; $\alpha \leftarrow \alpha - \nabla_{\alpha} \dfrac{1}{|B|}\displaystyle\sum_{\tau_{k} \in B} (\mathcal{H}^{\texttt{target}}  - \mathcal{H}(o)) \cdot \alpha$
\State Update target critic $\phi^{\texttt{target}}_{i} \leftarrow \rho \phi^{\texttt{target}}_{i} + (1-\rho)\phi_{i}$ for $i \in \{1, 2\}$
\EndFor
\EndFor
\end{algorithmic}
\end{algorithm*}

\section{Conclusion}

By combining FastSAC and FastTD3, scalable off-policy RL algorithms, with a streamlined training pipeline, our recipe closes the gap between the promise of high-throughput parallel simulation and the practical demands of sim-to-real humanoid learning.
In particular, we show that off-policy RL algorithms can be scaled effectively to reduce sim-to-real iteration time for learning whole-body humanoid controllers.
In this work, we have intentionally maintained a simple, minimalist design that other researchers can easily build upon.
We expect that incorporating recent advances in off-policy RL and humanoid learning into this recipe will push the state-of-the-art even further.
To support such progress, we provide an open-source implementation of our recipe \citep{Holosoma}.
We hope this report serves as a blueprint for researchers aiming to rapidly iterate on humanoid policies.

\newpage

\bibliography{reference}

@inproceedings{bellemare2017distributional,
  title     = {A distributional perspective on reinforcement learning},
  author    = {Bellemare, Marc G and Dabney, Will and Munos, R{\'e}mi},
  booktitle = {International Conference on Machine Learning},
  year      = {2017}
}

@inproceedings{schwarzer2023bigger,
  title     = {Bigger, better, faster: Human-level atari with human-level efficiency},
  author    = {Schwarzer, Max and Ceron, Johan Samir Obando and Courville, Aaron and Bellemare, Marc G and Agarwal, Rishabh and Castro, Pablo Samuel},
  booktitle = {International Conference on Machine Learning},
  year      = {2023}
}

@inproceedings{ball2023efficient,
  title     = {Efficient online reinforcement learning with offline data},
  author    = {Ball, Philip J and Smith, Laura and Kostrikov, Ilya and Levine, Sergey},
  booktitle = {International Conference on Machine Learning},
  year      = {2023}
}

@article{ba2016layer,
  title   = {Layer normalization},
  author  = {Ba, Jimmy Lei and Kiros, Jamie Ryan and Hinton, Geoffrey E},
  journal = {arXiv preprint arXiv:1607.06450},
  year    = {2016}
}

@inproceedings{dabney2018distributional,
  title     = {Distributional reinforcement learning with quantile regression},
  author    = {Dabney, Will and Rowland, Mark and Bellemare, Marc and Munos, R{\'e}mi},
  booktitle = {Proceedings of the AAAI conference on artificial intelligence},
  year      = {2018}
}

@article{akkaya2019solving,
  title   = {Solving rubik's cube with a robot hand},
  author  = {Akkaya, Ilge and Andrychowicz, Marcin and Chociej, Maciek and Litwin, Mateusz and McGrew, Bob and Petron, Arthur and Paino, Alex and Plappert, Matthias and Powell, Glenn and Ribas, Raphael and others},
  journal = {arXiv preprint arXiv:1910.07113},
  year    = {2019}
}

@inproceedings{fujimoto2018addressing,
  title     = {Addressing function approximation error in actor-critic methods},
  author    = {Fujimoto, Scott and Hoof, Herke and Meger, David},
  booktitle = {International Conference on Machine Learning},
  year      = {2018}
}

@article{schulman2017proximal,
  title   = {Proximal policy optimization algorithms},
  author  = {Schulman, John and Wolski, Filip and Dhariwal, Prafulla and Radford, Alec and Klimov, Oleg},
  journal = {arXiv preprint arXiv:1707.06347},
  year    = {2017}
}

@article{haarnoja2018soft2,
  title   = {Soft actor-critic algorithms and applications},
  author  = {Haarnoja, Tuomas and Zhou, Aurick and Hartikainen, Kristian and Tucker, George and Ha, Sehoon and Tan, Jie and Kumar, Vikash and Zhu, Henry and Gupta, Abhishek and Abbeel, Pieter and others},
  journal = {arXiv preprint arXiv:1812.05905},
  year    = {2018}
}

@inproceedings{agarwal2023legged,
  title     = {Legged locomotion in challenging terrains using egocentric vision},
  author    = {Agarwal, Ananye and Kumar, Ashish and Malik, Jitendra and Pathak, Deepak},
  booktitle = {Conference on robot learning},
  year      = {2023}
}

@article{haarnoja2018learning,
  title   = {Learning to walk via deep reinforcement learning},
  author  = {Haarnoja, Tuomas and Ha, Sehoon and Zhou, Aurick and Tan, Jie and Tucker, George and Levine, Sergey},
  journal = {arXiv preprint arXiv:1812.11103},
  year    = {2018}
}

@article{kaufmann2023champion,
  title={Champion-level drone racing using deep reinforcement learning},
  author={Kaufmann, Elia and Bauersfeld, Leonard and Loquercio, Antonio and M{\"u}ller, Matthias and Koltun, Vladlen and Scaramuzza, Davide},
  journal={Nature},
  volume={620},
  number={7976},
  pages={982--987},
  year={2023},
  publisher={Nature Publishing Group UK London}
}

@inproceedings{lee2025hyperspherical,
  title={Hyperspherical Normalization for Scalable Deep Reinforcement Learning},
  author={Lee, Hojoon and Lee, Youngdo and Seno, Takuma and Kim, Donghu and Stone, Peter and Choo, Jaegul},
  booktitle={International Conference on Machine Learning},
  year={2025}
}

@inproceedings{nauman2024bigger,
  title={Bigger, regularized, optimistic: scaling for compute and sample-efficient continuous control},
  author={Nauman, Michal and Ostaszewski, Mateusz and Jankowski, Krzysztof and Mi{\l}o{\'s}, Piotr and Cygan, Marek},
  booktitle={Advances in Neural Information Processing Systems},
  year={2024}
}

@inproceedings{lee2024simba,
  title={Simba: Simplicity bias for scaling up parameters in deep reinforcement learning},
  author={Lee, Hojoon and Hwang, Dongyoon and Kim, Donghu and Kim, Hyunseung and Tai, Jun Jet and Subramanian, Kaushik and Wurman, Peter R and Choo, Jaegul and Stone, Peter and Seno, Takuma},
  booktitle={International Conference on Learning Representations},
  year={2024}
}

@inproceedings{sukhija2024maxinforl,
  title={MaxInfoRL: Boosting exploration in reinforcement learning through information gain maximization},
  author={Sukhija, Bhavya and Coros, Stelian and Krause, Andreas and Abbeel, Pieter and Sferrazza, Carmelo},
  booktitle={International Conference on Learning Representations},
  year={2025}
}

@article{zakka2025mujoco,
  title={MuJoCo Playground},
  author={Zakka, Kevin and Tabanpour, Baruch and Liao, Qiayuan and Haiderbhai, Mustafa and Holt, Samuel and Luo, Jing Yuan and Allshire, Arthur and Frey, Erik and Sreenath, Koushil and Kahrs, Lueder A and others},
  journal={arXiv preprint arXiv:2502.08844},
  year={2025}
}

@article{mittal2023orbit,
   author={Mittal, Mayank and Yu, Calvin and Yu, Qinxi and Liu, Jingzhou and Rudin, Nikita and Hoeller, David and Yuan, Jia Lin and Singh, Ritvik and Guo, Yunrong and Mazhar, Hammad and Mandlekar, Ajay and Babich, Buck and State, Gavriel and Hutter, Marco and Garg, Animesh},
   journal={IEEE Robotics and Automation Letters},
   title={Orbit: A Unified Simulation Framework for Interactive Robot Learning Environments},
   year={2023},
   volume={8},
   number={6},
   pages={3740-3747},
   doi={10.1109/LRA.2023.3270034}
}

@inproceedings{li2023parallel,
  title={Parallel $ Q $-Learning: Scaling Off-policy Reinforcement Learning under Massively Parallel Simulation},
  author={Li, Zechu and Chen, Tao and Hong, Zhang-Wei and Ajay, Anurag and Agrawal, Pulkit},
  booktitle={International Conference on Machine Learning},
  pages={19440--19459},
  year={2023},
  organization={PMLR}
}

@InProceedings{rudin2022learning,
  title = 	 {Learning to Walk in Minutes Using Massively Parallel Deep Reinforcement Learning},
  author =       {Rudin, Nikita and Hoeller, David and Reist, Philipp and Hutter, Marco},
  booktitle = 	 {Proceedings of the 5th Conference on Robot Learning},
  pages = 	 {91--100},
  year = 	 {2022},
  volume = 	 {164},
  series = 	 {Proceedings of Machine Learning Research},
  publisher =    {PMLR},
  url = 	 {https://proceedings.mlr.press/v164/rudin22a.html},
}

@article{huang2022cleanrl,
  author  = {Shengyi Huang and Rousslan Fernand Julien Dossa and Chang Ye and Jeff Braga and Dipam Chakraborty and Kinal Mehta and João G.M. Araújo},
  title   = {CleanRL: High-quality Single-file Implementations of Deep Reinforcement Learning Algorithms},
  journal = {Journal of Machine Learning Research},
  year    = {2022},
  volume  = {23},
  number  = {274},
  pages   = {1--18},
  url     = {http://jmlr.org/papers/v23/21-1342.html}
}

@article{raffin2025isaacsim,
  title   = "Getting SAC to Work on a Massive Parallel Simulator: An RL Journey With Off-Policy Algorithms",
  author  = "Raffin, Antonin",
  journal = "araffin.github.io",
  year    = "2025",
  month   = "Feb",
  url     = "https://araffin.github.io/post/sac-massive-sim/"
}

@article{shukla2025fastsac,
  title   = "Speeding Up SAC with Massively Parallel Simulation",
  author  = "Shukla, Arth",
  journal = "https://arthshukla.substack.com",
  year    = "2025",
  month   = "Mar",
  url     = "https://arthshukla.substack.com/p/speeding-up-sac-with-massively-parallel"
}

@article{heess2017emergence,
  title={Emergence of locomotion behaviours in rich environments},
  author={Heess, Nicolas and Tb, Dhruva and Sriram, Srinivasan and Lemmon, Jay and Merel, Josh and Wayne, Greg and Tassa, Yuval and Erez, Tom and Wang, Ziyu and Eslami, SM and others},
  journal={arXiv preprint arXiv:1707.02286},
  year={2017}
}

@misc{makoviychuk2021isaac,
      title={Isaac Gym: High Performance GPU-Based Physics Simulation For Robot Learning}, 
      author={Viktor Makoviychuk and Lukasz Wawrzyniak and Yunrong Guo and Michelle Lu and Kier Storey and Miles Macklin and David Hoeller and Nikita Rudin and Arthur Allshire and Ankur Handa and Gavriel State},
      year={2021},
      journal={arXiv preprint arXiv:2108.10470}
}

@article{seo2025fasttd3,
  title={FastTD3: Simple, Fast, and Capable Reinforcement Learning for Humanoid Control},
  author={Seo, Younggyo and Sferrazza, Carmelo and Geng, Haoran and Nauman, Michal and Yin, Zhao-Heng and Abbeel, Pieter},
  journal={arXiv preprint arXiv:2505.22642},
  year={2025}
}

@inproceedings{kingma2014adam,
  title={Adam: A method for stochastic optimization},
  author={Kingma, Diederik P and Ba, Jimmy},
  booktitle={International Conference on Learning Representations},
  year={2015}
}

@inproceedings{zhai2023sigmoid,
  title={Sigmoid loss for language image pre-training},
  author={Zhai, Xiaohua and Mustafa, Basil and Kolesnikov, Alexander and Beyer, Lucas},
  booktitle={Proceedings of the IEEE/CVF international conference on computer vision},
  pages={11975--11986},
  year={2023}
}

@article{obando2025simplicial,
  title={Simplicial Embeddings Improve Sample Efficiency in Actor-Critic Agents},
  author={Obando-Ceron, Johan and Mayor, Walter and Lavoie, Samuel and Fujimoto, Scott and Courville, Aaron and Castro, Pablo Samuel},
  journal={arXiv preprint arXiv:2510.13704},
  year={2025}
}

@inproceedings{tobin2017domain,
  title={Domain randomization for transferring deep neural networks from simulation to the real world},
  author={Tobin, Josh and Fong, Rachel and Ray, Alex and Schneider, Jonas and Zaremba, Wojciech and Abbeel, Pieter},
  booktitle={2017 IEEE/RSJ international conference on intelligent robots and systems (IROS)},
  pages={23--30},
  year={2017},
  organization={IEEE}
}

@article{sadeghi2016cad2rl,
  title={Cad2rl: Real single-image flight without a single real image},
  author={Sadeghi, Fereshteh and Levine, Sergey},
  journal={arXiv preprint arXiv:1611.04201},
  year={2016}
}

@inproceedings{peng2018sim,
  title={Sim-to-real transfer of robotic control with dynamics randomization},
  author={Peng, Xue Bin and Andrychowicz, Marcin and Zaremba, Wojciech and Abbeel, Pieter},
  booktitle={2018 IEEE international conference on robotics and automation (ICRA)},
  pages={3803--3810},
  year={2018},
  organization={IEEE}
}

@inproceedings{phase_rewards,
  author       = {Yecheng Shao and
                  Yongbin Jin and
                  Xianwei Liu and
                  Weiyan He and
                  Hongtao Wang and
                  Wei Yang},
  title        = {Learning Free Gait Transition for Quadruped Robots via Phase-Guided
                  Controller},
  journal      = {CoRR},
  volume       = {abs/2201.00206},
  year         = {2022},
  url          = {https://arxiv.org/abs/2201.00206},
  eprinttype    = {arXiv},
  eprint       = {2201.00206},
  bibsource    = {dblp computer science bibliography, https://dblp.org}
}

@misc{HumanoidVerse,
  author = {CMU LeCAR Lab},
  title = {HumanoidVerse: A Multi-Simulator Framework for Humanoid Robot Sim-to-Real Learning},
  year = {2025},
  publisher = {GitHub},
  journal = {GitHub repository},
  howpublished = {\url{https://github.com/LeCAR-Lab/HumanoidVerse}},
}

@article{liao2025beyondmimic,
  title={Beyondmimic: From motion tracking to versatile humanoid control via guided diffusion},
  author={Liao, Qiayuan and Truong, Takara E and Huang, Xiaoyu and Tevet, Guy and Sreenath, Koushil and Liu, C Karen},
  journal={arXiv preprint arXiv:2508.08241},
  year={2025}
}

@article{peng2018deepmimic,
  title={Deepmimic: Example-guided deep reinforcement learning of physics-based character skills},
  author={Peng, Xue Bin and Abbeel, Pieter and Levine, Sergey and Van de Panne, Michiel},
  journal={ACM Transactions On Graphics (TOG)},
  year={2018},
}

@inproceedings{mittal2024symmetry,
  title={Symmetry considerations for learning task symmetric robot policies},
  author={Mittal, Mayank and Rudin, Nikita and Klemm, Victor and Allshire, Arthur and Hutter, Marco},
  booktitle={2024 IEEE International Conference on Robotics and Automation (ICRA)},
  pages={7433--7439},
  year={2024},
  organization={IEEE}
}

@article{stooke2018accelerated,
  title={Accelerated methods for deep reinforcement learning},
  author={Stooke, Adam and Abbeel, Pieter},
  journal={arXiv preprint arXiv:1803.02811},
  year={2018}
}

@inproceedings{liang2018gpu,
  title={Gpu-accelerated robotic simulation for distributed reinforcement learning},
  author={Liang, Jacky and Makoviychuk, Viktor and Handa, Ankur and Chentanez, Nuttapong and Macklin, Miles and Fox, Dieter},
  booktitle={Conference on Robot Learning},
  year={2018},
}

@inproceedings{espeholt2018impala,
  title={Impala: Scalable distributed deep-rl with importance weighted actor-learner architectures},
  author={Espeholt, Lasse and Soyer, Hubert and Munos, Remi and Simonyan, Karen and Mnih, Vlad and Ward, Tom and Doron, Yotam and Firoiu, Vlad and Harley, Tim and Dunning, Iain and others},
  booktitle={International conference on machine learning},
  year={2018},
}

@inproceedings{zhao2020sim,
  title={Sim-to-real transfer in deep reinforcement learning for robotics: a survey},
  author={Zhao, Wenshuai and Queralta, Jorge Pe{\~n}a and Westerlund, Tomi},
  booktitle={2020 IEEE symposium series on computational intelligence (SSCI)},
  year={2020},
}

@inproceedings{chebotar2019closing,
  title={Closing the sim-to-real loop: Adapting simulation randomization with real world experience},
  author={Chebotar, Yevgen and Handa, Ankur and Makoviychuk, Viktor and Macklin, Miles and Issac, Jan and Ratliff, Nathan and Fox, Dieter},
  booktitle={2019 International Conference on Robotics and Automation (ICRA)},
  year={2019},
}

@misc{Genesis,
  author = {Genesis Authors},
  title = {Genesis: A Generative and Universal Physics Engine for Robotics and Beyond},
  month = {December},
  year = {2024},
  url = {https://github.com/Genesis-Embodied-AI/Genesis}
}

@inproceedings{cheng2024extreme,
  title={Extreme parkour with legged robots},
  author={Cheng, Xuxin and Shi, Kexin and Agarwal, Ananye and Pathak, Deepak},
  booktitle={2024 IEEE International Conference on Robotics and Automation (ICRA)},
  year={2024},
}

@article{li2025reinforcement,
  title={Reinforcement learning for versatile, dynamic, and robust bipedal locomotion control},
  author={Li, Zhongyu and Peng, Xue Bin and Abbeel, Pieter and Levine, Sergey and Berseth, Glen and Sreenath, Koushil},
  journal={The International Journal of Robotics Research},
  year={2025},
}

@article{radosavovic2024learning,
  title={Learning humanoid locomotion over challenging terrain},
  author={Radosavovic, Ilija and Kamat, Sarthak and Darrell, Trevor and Malik, Jitendra},
  journal={arXiv preprint arXiv:2410.03654},
  year={2024}
}

@article{singh2024dextrah,
  title={Dextrah-rgb: Visuomotor policies to grasp anything with dexterous hands},
  author={Singh, Ritvik and Allshire, Arthur and Handa, Ankur and Ratliff, Nathan and Van Wyk, Karl},
  journal={arXiv preprint arXiv:2412.01791},
  year={2024}
}

@article{zhuang2024humanoid,
  title={Humanoid parkour learning},
  author={Zhuang, Ziwen and Yao, Shenzhe and Zhao, Hang},
  journal={arXiv preprint arXiv:2406.10759},
  year={2024}
}

@inproceedings{he2025hover,
  title={Hover: Versatile neural whole-body controller for humanoid robots},
  author={He, Tairan and Xiao, Wenli and Lin, Toru and Luo, Zhengyi and Xu, Zhenjia and Jiang, Zhenyu and Kautz, Jan and Liu, Changliu and Shi, Guanya and Wang, Xiaolong and others},
  booktitle={2025 IEEE International Conference on Robotics and Automation (ICRA)},
  year={2025},
}

@inproceedings{he2025asap,
  title={Asap: Aligning simulation and real-world physics for learning agile humanoid whole-body skills},
  author={He, Tairan and Gao, Jiawei and Xiao, Wenli and Zhang, Yuanhang and Wang, Zi and Wang, Jiashun and Luo, Zhengyi and He, Guanqi and Sobanbab, Nikhil and Pan, Chaoyi and others},
  booktitle={Robotics: Science and Systems},
  year={2025}
}

@article{schwarke2025rsl,
  title={Rsl-rl: A learning library for robotics research},
  author={Schwarke, Clemens and Mittal, Mayank and Rudin, Nikita and Hoeller, David and Hutter, Marco},
  journal={arXiv preprint arXiv:2509.10771},
  year={2025}
}

@misc{Holosoma,
  title = {Holosoma},
  author = {Amazon FAR and Abbeel, Pieter and Chen, Juyue and Duan, Rocky and Escontrela, Alejandro and Gandhi, Manan and Gundry, Samuel and Huang, Xiaoyu and Kanazawa, Angjoo and Lewicki, Tomasz and Li, Jiaman and Liu, Karen and Rosenthal, Clay and Seo, Younggyo and Sferrazza, Carlo and Shi, Guanya and Shih, Linda and Tseng, Jonathan and Wu, Zhen and Yang, Lujie and Yi, Brent and Zhang, Yuanhang},
  url = {https://github.com/amazon-far/holosoma},
  year = {2025}
}
\bibliographystyle{ref_bst}

\end{document}